\pdfoutput=1
\documentclass[11pt]{article}

\usepackage{ACL2023}

\usepackage{times}
\usepackage{latexsym}

\usepackage[T1]{fontenc}
\usepackage[utf8]{inputenc}

\usepackage{microtype}
\usepackage{amsmath}

\usepackage{inconsolata}

\usepackage{graphicx}
\usepackage[figuresright]{rotating}
\usepackage{multirow}
\title{LM-Cocktail: Resilient Tuning of Language Models via Model Merging}

\author{
Shitao Xiao$~^{\spadesuit}$ \ \ \ \ \
Zheng Liu$~^{\spadesuit}$\thanks{~~Correspondence author} \ \ \ \ \
Peitian Zhang$~^{\spadesuit}$ \ \ \ \ \ Xingrun Xing$~^{\clubsuit}$
\\ 
$^\spadesuit$ Beijing Academy of Artificial Intelligence \ \ \ \ \ \\
$^\clubsuit$ Institute of Automation, Chinese Academy of Sciences \ \ \ \ \ \\
{\tt stxiao@baai.ac.cn} \ \ \
{\tt \{zhengliu1026,namespace.pt\}@gmail.com} \\
{\tt xingxingrun2023@ia.ac.cn}
}

\begin{document}
\maketitle

\begin{abstract}
The pre-trained language models are continually fine-tuned to better support downstream applications. However, this operation may result in significant performance degeneration on general tasks beyond the targeted domain. To overcome this problem, we propose \textbf{LM-Cocktail} which enables the fine-tuned model to stay resilient in general perspectives. Our method is conducted in the form of model merging, where the fine-tuned language model is merged with the pre-trained base model or the peer models from other domains through weighted average. Despite simplicity, LM-Cocktail is surprisingly effective: the resulted model is able to achieve a strong empirical performance in the whole scope of general tasks while preserving a superior capacity in its targeted domain. 
% LM-Cocktail also enjoys a high versatility. It is applicable to both decoder-based LM on generation tasks and encoder-based LM on representation tasks. Meanwhile, it can well-conserve its competitiveness even in the absence of sufficient fine-tuning data or diverse peer models. 
We conduct comprehensive experiments with LLama and BGE models on popular benchmarks, including FLAN, MMLU, MTEB, whose results validate the efficacy of our proposed method. The code and checkpoints are available at https://github.com/FlagOpen/FlagEmbedding.
\end{abstract}

\section{Introduction}

\begin{figure*}[ht]
    \centering
    \scalebox{0.65}[0.65]
    {\includegraphics{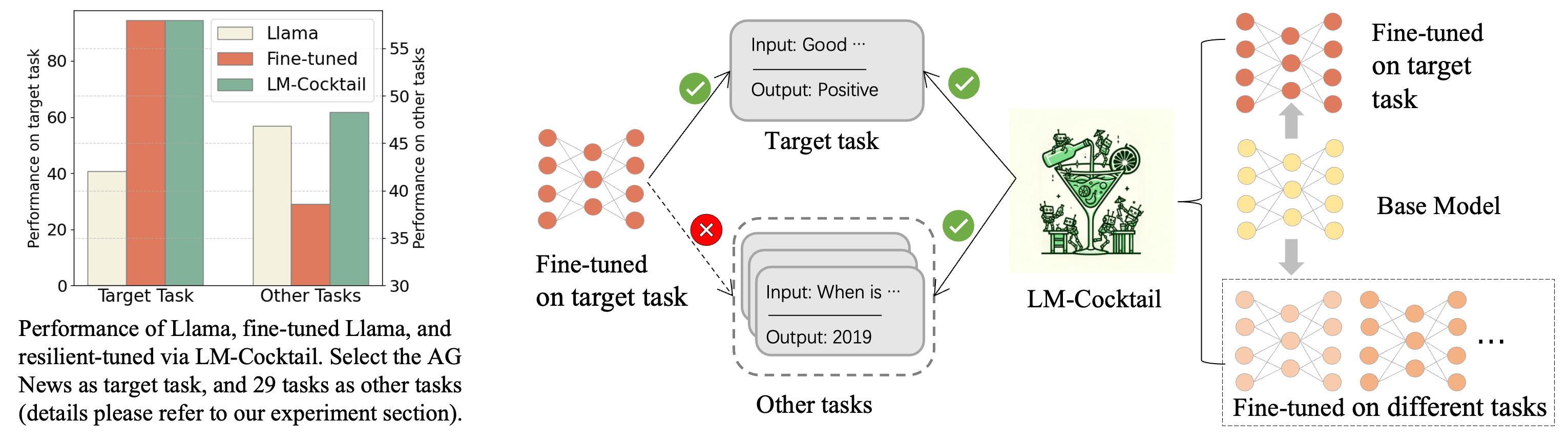}}
    \vspace{-5pt}
    \caption{The illustration of LM-Cocktail. Fine-tuning for the target task will
lead to severe degeneration of LM’s general
capabilities beyond the targeted domain. LM-Cocktail can increase accuracy on new target tasks while maintaining its accuracy on other tasks.}
    \label{fig:illustration}
    \vspace{-10pt}
\end{figure*}

% \begin{figure*}[!ht]
% \centering
% \subfigure{
% \scalebox{0.3}[0.3]{\includegraphics{figures/agnews.png}}
% }
% \subfigure{
% \scalebox{0.6}[0.6]{\includegraphics{figures/图片 1.png}}
% }
% \caption{Performance with different mixing weight.}
% \label{fig:alpha}
% \end{figure*}

Language models (LM) are fundamental pillars of artificial intelligence and natural language processing. Thanks to the considerable expansion of training scale and model size~\cite{devlin2018bert,liu2019roberta,raffel2020exploring,radford2019language,gpt3}, language models have made remarkable breakthroughs on a wide variety of NLP tasks, including representation, understanding, reasoning, and generation. In recent years, language models have been used as a crucial building block for many applications, such as information retrieval, conversational systems, and autonomous AI agents. In many of the applications, language models are frequently used via the ``pre-training and fine-tuning'' paradigm. Particularly, a generalist LM is pre-trained in the first place through an unsupervised or general-purpose supervised learning process \cite{gpt3,touvron2023llama,wei2022flan,wei2021finetuned,ouyang2022training}; then, the pre-trained generalist model is fine-tuned to be a specialist model for a down-stream task on top of certain in-domain data. 

Despite the improved performance in each particular application, the fine-tuning operation could lead to severe degeneration of LM's general capabilities beyond the targeted domain. Such a phenomenon is commonly referred as catastrophic forgetting \cite{goodfellow2013empirical,kirkpatrick2017overcoming,thompson2019overcoming,chen2020recall}.
As shown in Figure~\ref{fig:illustration}, fine-tuning Llama model on the target task can significantly improve its performance on the target task, but decrease its performance on other unrelated tasks.
In many real-world scenarios, catastrophic forgetting is unwelcome because language models need to exhibit both specialist and generalist characteristics simultaneously\cite{roziere2023code,chen2021evaluating,singhal2022large}. 
% For instance, the task-oriented language models are developed to provide user with domain-specific knowledge; meanwhile, they also are required to support user in terms of general functionalities, such as conversation, question answering, and common-sense reasoning \cite{roziere2023code,chen2021evaluating,singhal2022large}. 

The combat against catastrophic forgetting represents a sustained campaign within the machine learning communities, where numerous approaches have been continually proposed in recent years. There are two representative strategies which are widely adopted as the designing logic by many existing methods. One strategy is to rely on experience replay, where the model is learned with the mixed training data from both the new task and the previous tasks \cite{rolnick2019experience,shin2017continual}. The other strategy is to leverage regularization, where the changes in predictions or weights are regularized between the newly fine-tuned model and the historical pre-trained one \cite{kirkpatrick2017overcoming,li2017learning,rannen2017encoder}. However, it remains to explore more effective methods in the context of fine-tuned language models given the practical constraints of the existing methods. On one hand, it is infeasible to fully collect the training samples for all previous tasks, and have the model trained over again on the historical data once a new task is presented. On the other hand, the regularization may result in major changes to the existing fine-tuning operations, which could be incompatible with the well-established fine-tuning pipeline.  

In this work, we aim to design an effective framework to confront catastrophic forgetting, which will enable the fine-tuned language models to stay resilient in general tasks. Besides, we also expect the new framework to be more practical, which means it must be simple to conduct and fully compatible with the common model training workflow.  

With these considerations, we propose a new approach, called LM-Cocktail, which continually adapts well-fine-tuned language models on top of {m}odel {m}erging \cite{model-soups}. 
% The general form of LM-Cocktail is:
% \begin{equation}
%     \mathcal{M}_r = \alpha * \mathcal{M}_t + (1-\alpha)\sum_{\mathcal{M}_b, \{\mathcal{M}_d\}_{\mathcal{O}}} w_i \mathcal{M}_i, 
% \label{eqn:mix_pool}
% \end{equation}
% where $\mathcal{M}_t$ is the fine-tuned model, $\mathcal{M}_d$ is the pre-trained base model, and $\{\mathcal{M}_d\}_{\mathcal{O}}$ is a set of models fine-tuned on other domains. By LM-Cocktail, we can produce a resilient-tuned model $\mathcal{M}_r$, which has higher accuracy in the target task but maintains competitive performance on general tasks.
LM-Cocktail is a general paradigm, which can work under several different conditions. In the simplest form, it directly merges the fine-tuned model with the pre-trained base model to improve the general capabilities of the fine-tuned model. It can further accommodate more peer models fine-tuned for other general domains, and result in stronger empirical performances on top of merging weights estimated by few-shot validation examples. Finally, even at the absence of fine-tuning data, the merging strategy can be still applied to the remaining pre-trained base model and the fine-tuned models in other general domains for a competitive resilience.

% It merges the fine-tuned model with a diverse group of peer models in the form of weighted average, whereby the domain specialty from the fine-tuned model and the generality of the diverse peer models can be jointly acquired. There are two major concerns regarding the conduct of model mixing: 1) which group of models to mix, 2) how to determine the mixing weights. In our work, we introduce the following paradigm as a general solution. For one thing, model group is made up of three compositions: the pre-trained model, the fine-tuned model for the target domain, the fine-tuned models in other domains. For an other thing, the merging weights are directly computed from each model's prediction loss on the few-shot examples of the target domain. 

Our proposed method leads to a couple of immediate advantages given its working mechanism. First of all, LM-Cocktail is extremely \textit{simple}: the mixing weights can be directly derived from validation samples where no expensive training operations are needed. Secondly, LM-Cocktail is fully \textit{compatible} with the existing training pipeline, knowing that it simply works as a post-refinement step following the fine-tuning process. Above all, LM-Cocktail is \textit{empirically competitive}. According to our evaluations on three representative benchmarks, including FLAN \cite{wei2021finetuned}, MMLU \cite{hendrycks2020measuring}, and MTEB \cite{muennighoff2022mteb}, LM-Cocktail achieves a strong resilience in general domain tasks while preserving a superior fine-tuning performance on its targeted domain. Finally, LM-Cocktail turns out to be universally applicable: it can substantially contribute to both the decoder-based LM in language generation tasks and the encoder-based LM in language representation tasks. 

% To summarize, the contributions are three-fold: 1) technical design, a simple, effective, and practical method to adapt the fine-tuned model for high resilience, 2) comprehensive evaluations, comprehensively verify the empirical competitiveness for different types of models under diverse settings. 

% \section{Related Works}
% % related works: language model application and fine-tuning, catastrophic forgetting
% % related works: catastrophic forgetting and how to mitigate this problem
% % related works: model soup, ensemble, data privacy, personalization 

\section{LM-Cocktail}

\subsection{General Paradigm}

As a prerequisite condition, we are given a base language model, denoted as $\mathcal{M}_b$, which are well pre-trained for general applications. Typical examples of the base model can be Llama-Chat \cite{touvron2023llama} and FLAN-PaLM \cite{chung2022scaling}, which are LLMs learned from massive unsupervised learning and general-purse supervised fine-tuning. The base LM is continually fine-tuned to support one targeted down-stream task ($t$) with domain-specific training samples $\boldsymbol{X}_t$, which results in the fine-tuned model for the corresponding task: $\mathcal{M}_t$.

However, 
the fine-tuned model $\mathcal{M}_t$
is prone to degenerate empirical performances (catastrophic forgetting) on other general domains beyond the targeted domain $t$. The goal of LM-Cocktail is to maintain the general capabilities when fine-tuning on the target task. 
The core of \textbf{LM-Cocktail} is combining multiple models (with the same architecture but different weights) into a unified one by aggregating the weights from different models.
In this way, the resilient fine-tuned model can integrate the strengths from multiple individual models.

% Model merging is a specific way of model ensemble. It combines multiple models (with the same architecture but different weights) into a unified one by aggregating the weights from different models. It possesses the merits of ensemble learning, which integrates the strengths from multiple individual models. Meanwhile, it is a more computationally efficient paradigm. Because of these features, it is promising to realize the resilient fine-tuned LM on top of model merging. Such an operation is named as \textbf{LM-Cocktail} in our work. 

To derive the appropriate model merging strategy for LM-Cocktail, there are two fundamental problems to solve: 1) which group of candidate models to merge, 2) how to determine the merging weights. Knowing that the resilient fine-tuned LM is to restore the degenerated performances in general domains, there are two sources of candidate models to consider. One source is the pre-trained base model $\mathcal{M}_b$, the other source is the entire group of fine-tuned models in other domains ($\{\mathcal{M}_d\}_{\mathcal{D}}$). Without loss of generality, we derive the following form of merging function: 
\begin{equation}
    \mathcal{M}_r \leftarrow
    \alpha \mathcal{M}_t + (1-\alpha) \sum_{\mathcal{M}_b,  \{\mathcal{M}_d\}_{\mathcal{D}}} w_i * \mathcal{M}_i,
    % ~~ \mathcal{M}_i \in \big{\{} \mathcal{M}_b, \mathcal{M}_t, \{\mathcal{M}_d\}_{\mathcal{O}} \big{\}}. 
\label{eqn:base}
\end{equation}
where $\mathcal{M}_r$ is the resilient-tuned model, $\alpha$ is a hyper-parameter whose default value is 0.5, and $w_i$ indicates the merging weight which has been normalized: $\sum_i w_i = 1$. 
% In previous works, diverse forms of merging weights were designed for different applications, such as average merging \cite{model-soups}, Fisher merging \cite{matena2022merging}, and RegMean merging \cite{jin2022dataless}. 
For our case, we require the resilient-tuned model to preserve strong capacity as the directly fine-tuned model in its targeted domain while improving the general domain performance. Therefore, the candidate models' performances in the targeted domain are the critical indicators of merging weights. Based on this intuition, we introduce the following form of weight computation:
\begin{equation}
    w_i \leftarrow \mathrm{softmax}(-\mathcal{L}(\mathcal{M}_i, E_t) / \tau). 
\label{eqn:weight}
\end{equation}
In this function, $\mathcal{L}(\mathcal{M}_i, E_t)$ stands for the prediction loss of candidate model $\mathcal{M}_i$ on the few-shot examples $E_t$ from the targeted domain $t$, $\tau$ is the temperature to control the smoothness. That is to say, the larger loss on the targeted domain, the smaller weight is allocated to the candidate model. So we can
give lower coefficients to models that perform very badly in the target task.
The few-shot examples are a tiny group of hold-back samples from the targeted domain. According to our empirical study, 5-shot examples have been sufficiently competitive throughout different settings.

\subsection{Variations}
% mono-specialist variation
% few-shot variaion

The general form of LM-Cocktail in Eq \ref{eqn:base} requires the presence of three elements: the base model $\mathcal{M}_b$, the fine-tuned model for the targeted domain $\mathcal{M}_t$, and the fine-tuned models in other general domains $\{\mathcal{M}_d\}_\mathcal{D}$. Nevertheless, the general requirement can be largely relaxed to accommodate different real-world settings. Here, we introduce two common variational forms to confront the situations where either diverse general-domain specialists or targeted domain fine-tuning is not available. 

$\bullet$ \textbf{Mono-Specialist}. When the diverse fine-tuned models in general domains are absent, the merging function is simplified as the combination of base model $\mathcal{M}_b$ and the mono-specialist model from the targeted domain $\mathcal{M}_t$: 
\begin{equation}
    \mathcal{M}_r \leftarrow \alpha\mathcal{M}_t + (1-\alpha)\mathcal{M}_b.
\end{equation} 
% in which the merging parameter $\alpha$ computed as: 
% \begin{equation*}
%     \alpha \leftarrow \frac{\exp(\mathcal{L}(\mathcal{M}_t, E_t) / \tau)}{\exp(\mathcal{L}(\mathcal{M}_t, E_t) / \tau) + \exp(\mathcal{L}(\mathcal{M}_b, E_t) / \tau)}. 
% \end{equation*} 
Given that fine-tuned model $\mathcal{M}_t$ typically exhibits significantly lower loss compared to other models, we did not employ Eqn~\ref{eqn:weight} to calculate weights; instead, we introduce a hyperparameter $\alpha$. Experimental results demonstrate that simply setting $\alpha$ to 0.5 yields promising outcomes.

$\bullet$ \textbf{Without Fine-tuning}. The fine-tuning in the targeted domain can be constrained due to the absence of domain-specific data or computation resources. In this situation, the merging function is transformed into the combination of base model and fine-tuned model from general domains:
\begin{equation*}
    \mathcal{M}_r \leftarrow \sum_{\mathcal{M}_b, \{\mathcal{M}_d\}_{\mathcal{D}}} w_i * \mathcal{M}_i.
\end{equation*}
In this place, we assume the few-shot examples $E_t$ for merging weights (Eq. \ref{eqn:weight}) are always available, which is a very moderate condition in practice. In this manner, we obviate the need for training any new models; instead, by incurring minimal costs, we can seamlessly integrate existing models to obtain a model tailored for downstream tasks.

\begin{table*}[ht]
    \centering
    \scalebox{1.0}{
    % \footnotesize
    \begin{tabular}
    {llcccc}
    \hline
Fine-tune on & Performance on &  Llama & Fine-tuned & LM-Cocktail$_{2}$ & LM-Cocktail$_{10}$ \\ 
\hline 
\multirow{2}{*}{AG News} & AG News & 40.80 & 94.42 & {94.46} & 94.41\\ 
 & Other tasks  & 46.80 & 38.58 & 47.73 & {48.32} \\ 
\hline 
\multirow{2}{*}{Common Gen} & Common Gen & 21.14 & 39.20 & 41.22 & {41.45} \\ 
 & Other tasks  & 47.48 & 46.90 & 50.88 & {58.57} \\ 
\hline 
\multirow{2}{*}{MNLI} & MNLI & 32.14 & 87.90 & 88.88 & {89.23} \\ 
 & Other tasks  & 47.10 & 47.49 & 53.53 & {56.31} \\ 
\hline 
\multirow{2}{*}{Winogrande} & Winogrande & 60.93 & 75.45 & {77.90} & 77.03\\ 
 & Other tasks  & 46.11 & 47.33 & 50.52 & {58.52} \\ 
 \hline
\multirow{2}{*}{MRPC} & MRPC & 31.86 & {85.78} & 73.77 & 80.88\\ 
 & Other tasks  & {47.11} & 36.45 & 39.56 & 42.77\\ 
\hline 
\multirow{2}{*}{NQ} & NQ & 0.00 & 29.09 & 29.25 & {29.64} \\ 
 & Other tasks  & 48.21 & 52.19 & 54.58 & {60.28} \\ 
\hline 
\multirow{2}{*}{SQuAD} & SQuAD & 0.06 & 86.77 & 85.67 & {86.94} \\ 
 & Other tasks  & 48.21 & 49.48 & 51.64 & {54.09} \\ 
\hline 
\multirow{2}{*}{SST2} & SST2 & 63.30 & 95.53 & {96.56} & {96.56} \\ 
 & Other tasks  & {46.02} & 38.94 & 41.63 & 45.03\\ 
\hline 
\multirow{2}{*}{Hellaswag} & Hellaswag & 71.58 & 77.20 & {79.00} & 78.61\\ 
 & Other tasks  & 45.74 & 46.10 & 48.95 & {57.87} \\ 
\hline 
    \end{tabular}}
    % \vspace{-10pt}
    \caption{
    A comparative performance analysis between base model Llama, fine-tuned model, and resilient-tuned model via LM-Cocktail. LM-Cocktail$_{2}$ is produced by merging the base model and fine-tuned model, while LM-Cocktail$_{10}$ merges fine-tuned model, base model, and other models fine-tuned on 8 different tasks. 
There are a total of 30 test tasks, and "Others tasks" refers to the remaining 29 tasks after the corresponding task is removed.
    }
    \label{tab:llm-finetune-pool}
    \vspace{-10pt}
\end{table*}

\section{Experimental setup}
 
We conducted experiments with two types of models: decoder-based LM and encoder-based LM.
We fine-tuned 9 encoder-based models and 9 decoder-based models separately, and then evaluated the performance of fine-tuned models and resilient-tuned models. Following are the detailed experimental settings.

\subsection{Decoder-based LM}
\label{sec:llm_exp}
$\bullet$ \textbf{Base Model}. 
We use Llama-2-chat-7b~\footnote{https://huggingface.co/meta-llama/Llama-2-7b-chat-hf}~\cite{touvron2023llama}
as the base model, which has an impressive zero-shot ability on various tasks.

$\bullet$ \textbf{Fine-tune}. 
We use the datasets collected by~\cite{cheng2023uprise,llm_retriever}, which consist of 30 tasks from FLAN~\cite{wei2022flan}.
We select 9 different tasks from it to fine-tune the base model, including NQ, SQuAD, Hellaswag, SST2, Winogrande, CommonGen, MRPC, AG News, and MNLI. 
For more information of training data please refer to Appendix~\ref{sec:data_appendix}. 
The fine-tuned code is based on FastChat package\footnote{https://github.com/lm-sys/FastChat}. The learning rate is 2e-5, the batch size is 128, and the max number of epochs is 3. 
% For our method, the $\alpha$ defaults to 0.5 and the temperature is 5.

$\bullet$ \textbf{Evaluation}.
We evaluate the performance on the test set of 30 tasks collected by~\cite{cheng2023uprise,llm_retriever}.
The test data for fine-tuning tasks (NQ, SQuAD, Hellaswag, SST2, Winogrande, CommonGen, MRPC, AG News, and MNLI) are also included in this collection.
% There are 9 task clusters in this datasets, including Reading Comprehension, Closed-book QA, Paraphrase Detection,
% Natural Language Inference, Sentiment Analysis,
% Commonsense Reasoning, Coreference Resolution,
% Structure to Text, and Summarization.
The detailed metric for each task can refer to ~\cite{llm_retriever}.
Besides, we also conduct experiments on additional tasks from MMLU datasets, which is a widely used benchmark for LLMs.

\subsection{Encoder-based LM}
\label{sec:emb_exp}
$\bullet$ \textbf{Base Model}. We choose the bge-base-v$1.5$ embedding model~\footnote{https://huggingface.co/BAAI/bge-base-en-v1.5}~\cite{bge_embedding} as the base model for embedding tasks, which can map text into embedding representation.

$\bullet$ \textbf{Fine-tune}. We select 9 datasets from sentence transformers repo\footnote{https://huggingface.co/datasets/sentence-transformers/embedding-training-data}, including GooAQ, Yahoo Answers, MSMarco, Stack Exchange, ELI5, SQuAD, AmazonQA, Quora, HotpotQA. 
Appendix~\ref{sec:data_appendix} shows the details of training data. 
We fine-tune BGE model on these datasets with FlagEmbedding tool\footnote{https://github.com/FlagOpen/FlagEmbedding}. We use the AdamW optimizer with a learning rate of 2e-5. The batch size is 256, and the temperature for contrastive learning is 0.02. 
% For hyper-parameters in LM-Cocktail, the $\alpha$ defaults to 0.5 and the temperature is 5.

$\bullet$ \textbf{Evaluation}.
We evaluate the models with the 15 retrieval tasks in mteb benchmark~\cite{muennighoff2022mteb}, and use NDCG@10 as the evaluation metric. The test data of 3 fine-tuning tasks: MSMarco, HotpotQA and Quora are included in this benchmark.
For the purpose of facilitating training and testing across various tasks, we don't add the default query instruction from~\cite{bge_embedding}.

\begin{table*}[!t]
    \centering
    \scalebox{1.0}{
    % \footnotesize
    \begin{tabular}
    {llcccc}
    \hline
Fine-tune on & Performance on &  BGE & Fine-tuned & LM-Cocktail$_2$ & LM-Cocktail$_{10}$ \\ 
\hline 
\multirow{2}{*}{HotpotQA} & HotpotQA & 71.81 & 75.96 & 74.78 & 74.67 \\ 
 & Other tasks  & 49.81 & 47.49 & 49.98 & 50.64 \\ 
\hline 
\multirow{2}{*}{Quora} & Quora & 88.90 & 90.31 & 89.93 & 89.81 \\ 
 & Other tasks  & 48.59 & 47.43 & 48.09 & 49.11 \\ 
\hline 
\multirow{2}{*}{MSMarco} & MSMarco & 41.15 & 42.23 & 42.01 & 41.88 \\ 
 & Other tasks  & 52.00 & 51.98 & 52.71 & 53.22 \\ 
\hline 
    \end{tabular}}
    \vspace{-10pt}
    \caption{A comparative performance analysis between base model BGE, fine-tuned model, and resilient-tuned model via LM-Cocktail. LM-Cocktail$_{2}$ is produced by merging the base model and fine-tuned model, while LM-Cocktail$_{10}$ merges fine-tuned model, base model, and other models fine-tuned on 8 different tasks. 
There are a total of 15 test tasks, and "Others tasks" refers to the remaining 14 tasks after the corresponding task is removed.}
    \label{tab:emb-finetune-pool}
    \vspace{-10pt}
\end{table*}

\section{Experimental Results}

In this section, we show the experimental results and represent the key findings.
Firstly, we compare the performance of fine-tuned models and resilient-tuned models.
Next, we evaluate the performance of LM-Cocktail when fine-tuning on target task is unavailable.
Finally, we investigate the impact of weight $\alpha$ and the number of examples.

\subsection{Overall Comparison}

Our experiments compare the performance of base models, corresponding fine-tuned models, and models
resilient-tuned via LM-Cocktail. 
For each fine-tuned model, we measure its performance on the specific target task as well as its performance on other tasks. 
We also tested models resilient-tuned using our method, which include two variants: (1) \textbf{LM-Cocktail$_2$}: merge the fine-tuned model with the base model; (2) \textbf{LM-Cocktail$_{10}$}: merge 10 models, including model fine-tuned on target task, base model, other the remaining eight fine-tuned models from section~\ref{sec:llm_exp}.
We have summarized the results in Table~\ref{tab:llm-finetune-pool} and~\ref{tab:emb-finetune-pool}.
For detailed results for each test task please refer to Appendix~\ref{sec:results_appendix}.

\subsubsection{Analysis on decoder-based LM} 
\label{sec:ana_decoder}

% The terms AG News, Commen Gen, SQuAD, and Winogrande in the first line respectively denote models fine-tuned on AG News, Commen Gen, SQuAD, and Winogrande datasets. The notation ``+LM-Cocktail''
% indicates the resilient-tuned model that merged according to Eqn~\ref{eqn:base}. 
% Take  ``AG News + LM-Cocktail'' as an example: 
% We merge the model fine-tuned on AG News, the base model, and 8 models fine-tuned on other tasks (see section~\ref{sec:llm_exp} for all fine-tuning tasks).
% For more results please refer to Table~\ref{tab:llm-finetune-pool-appendix} in Appendix~\ref{appendix:results}.

From Table~\ref{tab:llm-finetune-pool}, we have following observations: 
(1) the fine-tuned model demonstrates significant improvement over the base model in the corresponding task. For example, the model fine-tuned on AG News achieves an accuracy of 94.42\% in the corresponding task, whereas the base model only achieves 40.9\% accuracy on the same task. 
(2) However, this gain comes at a cost: in other tasks, the fine-tuned model often lags behind the performance of the base model. For example, the accuracy of the fine-tuned model on other tasks is only 38.58\%, substantially lower than the 46.8\% accuracy of the base model. 
(3) In contrast, LM-Cocktail$_2$ maintains effectiveness in its corresponding task (94.46\% in AG News task) while also demonstrating competitive performance in other tasks (47.73\%). 
And LM-Cocktail$_{10}$ further enhances the performance of other tasks (the accuracy increases from 38.58\% to 48.32\% after merging). In most of the cases, LM-Cocktail$_2$ and LM-Cocktail$_{10}$ even outperform the base model on other tasks.
This finding demonstrates that our approach can integrate the strengths of models to be merged, and even surpass them in performance.
(4) Besides, 
fine-tuning on some tasks (e.g., NQ) can enhance performance not only on the corresponding task but also on other tasks; our proposed method remains effective on these tasks: LM-Cocktail achieves higher accuracy both in target task and other tasks. These findings demonstrate the versatility of our approach.

\begin{table*}[ht]
    \centering
    \scalebox{0.7}{
    \footnotesize
    \begin{tabular}
    % {|c||cc||c||cc||cc|}
    {cccccccc}
    \hline
 Dataset & Llama & Llama-ICL & Multitask-learning & LM-Cocktail$_{\textbf{blackbox}}$ & LM-Cocktail & LM-Cocktail'$^{u}_{\textbf{blackbox}}$ & LM-Cocktail$^{u}$ \\ 
 \hline
Avg & 45.87 & 46.65 & 32.88 & 42.28 & 48.01 & 47.46 & 48.21 \\ 
\hline
abstract-algebra & 28.0 & 30.0 & 21.0 & 29.0 & 35.0 & 33.0 & 34.0 \\ 
anatomy & 42.96 & 42.22 & 34.07 & 45.19 & 46.67 & 46.67 & 48.15 \\ 
astronomy & 44.08 & 48.03 & 34.21 & 46.05 & 46.05 & 44.08 & 47.37 \\ 
business-ethics & 42.0 & 42.0 & 41.0 & 50.0 & 46.0 & 52.0 & 48.0 \\ 
clinical-knowledge & 50.57 & 51.32 & 39.62 & 47.92 & 51.32 & 51.7 & 51.32 \\ 
college-biology & 50.0 & 52.78 & 27.08 & 41.67 & 52.08 & 49.31 & 51.39 \\ 
college-chemistry & 23.0 & 26.0 & 31.0 & 19.0 & 29.0 & 29.0 & 29.0 \\ 
college-computer-science & 29.0 & 37.0 & 37.0 & 33.0 & 46.0 & 43.0 & 45.0 \\ 
college-mathematics & 29.0 & 33.0 & 36.0 & 29.0 & 31.0 & 35.0 & 31.0 \\ 
college-medicine & 38.15 & 40.46 & 31.79 & 28.32 & 40.46 & 39.31 & 40.46 \\ 
college-physics & 21.57 & 24.51 & 20.59 & 21.57 & 19.61 & 19.61 & 19.61 \\ 
computer-security & 59.0 & 54.0 & 40.0 & 59.0 & 55.0 & 49.0 & 57.0 \\ 
conceptual-physics & 38.3 & 38.72 & 29.79 & 38.72 & 39.57 & 39.57 & 40.0 \\ 
econometrics & 28.95 & 33.33 & 22.81 & 28.07 & 26.32 & 35.09 & 28.07 \\ 
electrical-engineering & 42.76 & 43.45 & 34.48 & 33.1 & 46.9 & 44.14 & 47.59 \\ 
elementary-mathematics & 27.25 & 28.04 & 21.16 & 28.84 & 26.46 & 26.72 & 26.72 \\ 
formal-logic & 22.22 & 25.4 & 35.71 & 28.57 & 25.4 & 24.6 & 25.4 \\ 
global-facts & 41.0 & 31.0 & 26.0 & 37.0 & 35.0 & 33.0 & 35.0 \\ 
high-school-biology & 46.45 & 52.58 & 33.23 & 35.16 & 53.87 & 53.55 & 53.87 \\ 
high-school-chemistry & 30.54 & 33.99 & 21.67 & 30.05 & 32.02 & 32.51 & 31.53 \\ 
high-school-computer-science & 39.0 & 46.0 & 30.0 & 37.0 & 40.0 & 41.0 & 40.0 \\ 
high-school-european-history & 60.61 & 56.97 & 32.12 & 51.52 & 63.03 & 63.64 & 64.85 \\ 
high-school-geography & 57.07 & 59.6 & 33.33 & 55.56 & 60.1 & 61.62 & 59.09 \\ 
high-school-government-and-politics & 70.47 & 67.36 & 38.34 & 39.38 & 70.98 & 66.84 & 70.98 \\ 
high-school-macroeconomics & 39.23 & 41.54 & 31.79 & 38.21 & 45.64 & 44.1 & 44.87 \\ 
high-school-mathematics & 26.3 & 23.7 & 22.96 & 25.19 & 24.81 & 24.44 & 24.81 \\ 
high-school-microeconomics & 36.55 & 43.7 & 31.93 & 36.55 & 41.6 & 40.34 & 41.6 \\ 
high-school-physics & 25.83 & 28.48 & 29.8 & 27.81 & 29.14 & 30.46 & 29.14 \\ 
high-school-psychology & 59.63 & 64.04 & 33.94 & 60.0 & 65.32 & 65.87 & 64.77 \\ 
high-school-statistics & 23.61 & 31.48 & 36.57 & 25.46 & 28.24 & 30.09 & 28.7 \\ 
high-school-us-history & 65.2 & 66.18 & 38.24 & 65.2 & 65.69 & 66.67 & 64.71 \\ 
high-school-world-history & 61.18 & 66.24 & 35.02 & 57.81 & 66.24 & 65.82 & 65.4 \\ 
human-aging & 58.74 & 57.4 & 36.77 & 55.61 & 58.74 & 54.26 & 58.74 \\ 
human-sexuality & 54.96 & 48.09 & 38.17 & 29.01 & 57.25 & 55.73 & 56.49 \\ 
international-law & 59.5 & 57.02 & 42.15 & 48.76 & 61.16 & 58.68 & 60.33 \\ 
jurisprudence & 55.56 & 57.41 & 34.26 & 52.78 & 49.07 & 50.93 & 50.93 \\ 
logical-fallacies & 58.28 & 53.99 & 29.45 & 57.67 & 55.21 & 55.83 & 54.6 \\ 
machine-learning & 36.61 & 35.71 & 28.57 & 33.04 & 39.29 & 41.07 & 41.96 \\ 
management & 64.08 & 67.96 & 43.69 & 61.17 & 68.93 & 66.99 & 68.93 \\ 
marketing & 73.5 & 74.36 & 48.29 & 73.5 & 76.5 & 70.94 & 76.07 \\ 
medical-genetics & 46.0 & 53.0 & 32.0 & 49.0 & 50.0 & 51.0 & 49.0 \\ 
miscellaneous & 66.16 & 66.54 & 38.19 & 65.13 & 68.58 & 65.52 & 69.6 \\ 
moral-disputes & 50.87 & 52.89 & 28.32 & 44.22 & 49.42 & 49.13 & 50.0 \\ 
moral-scenarios & 24.25 & 21.34 & 24.8 & 23.35 & 24.25 & 24.25 & 24.25 \\ 
nutrition & 50.0 & 51.96 & 40.85 & 47.71 & 54.58 & 50.0 & 54.9 \\ 
philosophy & 51.77 & 56.91 & 30.87 & 47.59 & 54.02 & 53.7 & 53.05 \\ 
prehistory & 51.85 & 56.79 & 31.79 & 51.85 & 51.85 & 49.38 & 51.23 \\ 
professional-accounting & 34.75 & 35.46 & 29.08 & 34.75 & 37.23 & 36.52 & 37.94 \\ 
professional-law & 34.55 & 33.31 & 27.71 & 31.03 & 36.11 & 36.31 & 36.05 \\ 
professional-medicine & 40.44 & 34.19 & 29.41 & 25.37 & 43.01 & 44.85 & 43.75 \\ 
professional-psychology & 44.93 & 47.55 & 29.74 & 42.32 & 45.92 & 44.61 & 45.59 \\ 
public-relations & 53.64 & 51.82 & 30.0 & 38.18 & 56.36 & 54.55 & 56.36 \\ 
security-studies & 49.8 & 45.71 & 27.76 & 48.57 & 55.1 & 57.14 & 56.33 \\ 
sociology & 71.14 & 57.21 & 46.27 & 47.76 & 71.64 & 74.63 & 72.14 \\ 
us-foreign-policy & 73.0 & 68.0 & 42.0 & 67.0 & 71.0 & 67.0 & 73.0 \\ 
virology & 45.78 & 43.37 & 34.34 & 43.98 & 46.39 & 42.77 & 46.99 \\ 
world-religions & 64.91 & 67.84 & 37.43 & 61.99 & 70.18 & 67.84 & 70.18 \\ 
\hline
    \end{tabular}
    }
    \caption{Results of merging decoder models from other tasks. }
    \label{tab:llm_mix_mmlu}
    \vspace{-10pt}
\end{table*}

\subsubsection{Analysis on encoder-based LM}
The results of encoder models are shown in Table~\ref{tab:emb-finetune-pool}. We can observe the same trend in the section~\ref{sec:ana_decoder}:
The fine-tuned model achieves significant improvement over the base model in the corresponding task but has a lower accuracy on other unrelated tasks. LM-Cocktail$_2$ significantly enhances performance in downstream tasks while maintaining performance in other unrelated tasks. LM-Cocktail$_{10}$ further improves the general ability by merging the models fine-tuned on different tasks.
These results show the applicability of LM-Cocktail for both generative models and representation models, 
validating the universality of our proposed methodology.

\begin{table*}[ht]
    \centering
    \scalebox{0.8}{
    \footnotesize
    \begin{tabular}
    % {|c||c||c||cc||cc|}
    {ccccccc}
    \hline
Dataset & BGE &  Multitask-learning & LM-Cocktail$_{blackbox}$ & LM-Cocktail & LM-Cocktail'$^{u}_{blackbox}$ & LM-Cocktail$^{u}$ \\  
\hline
ArguAna & 63.61 & 59.18 & 61.43 & 64.34 & 65.07 & 64.31 \\ 
ClimateFEVER & 29.51 & 25.8 & 10.65 & 29.5 & 29.94 & 29.17 \\ 
DBPedia & 40.56 & 39.77 & 21.13 & 40.37 & 40.46 & 40.83 \\ 
FEVER & 83.66 & 73.76 & 83.91 & 86.07 & 84.21 & 86.1 \\ 
FiQA2018 & 39.11 & 41.7 & 38.53 & 41.89 & 39.2 & 42.05 \\ 
NFCorpus & 36.83 & 37.49 & 36.99 & 37.66 & 36.7 & 37.64 \\ 
NQ & 51.05 & 53.25 & 50.95 & 53.47 & 50.75 & 53.4 \\ 
SCIDOCS & 21.48 & 21.04 & 21.87 & 22.55 & 21.95 & 22.31 \\ 
SciFact & 73.81 & 73.82 & 74.31 & 75.14 & 73.31 & 75.12 \\ 
Touche2020 & 19.54 & 22.96 & 19.31 & 20.9 & 20.56 & 20.54 \\ 
TRECCOVID & 67.18 & 74.51 & 71.73 & 71.19 & 71.3 & 70.32 \\ 
CQADupstack & 41.04 & 42.74 & 39.72 & 43.03 & 40.9 & 43.03 \\ 
\hline
Avg & 47.28 & 47.17 & 44.21 & 48.84 & 47.86 & 48.73 \\ 
\hline
    \end{tabular}}
    \caption{Results of merging decoder models from other tasks.}
    \label{tab:emb_mix}
    \vspace{-10pt}
\end{table*}

\subsection{LM-Cocktail without Fine-tuning}
\label{sec:wo_finetune}
In many scenarios, fine-tuning on the target domain is not always available. For example, if there is not enough training dataset for a new task, fine-tuning a specialist model specific to this task is unfeasible. 
Besides, fine-tuning a separate model for each task is costly and inflexible, especially when fine-tuning large language models. We report the performance of LM-Cocktail without fine-tuning in Table~\ref{tab:llm_mix_mmlu} and~\ref{tab:emb_mix}

\subsubsection{Analysis on Decoder-based LM}
\label{sec:unseen_llm}
To evaluate the performance on tasks which haven't been seen in fine-tuning, we introduce additional tasks from MMLU benchmark. 
There are 57 tasks in MMLU, which are different from the fine-tuning tasks in section~\ref{sec:llm_exp}. We use the evaluation script and five-shot examples from the widely used framework EleutherAI~\footnote{https://github.com/EleutherAI/lm-evaluation-harness}. 

The results are summarized in Table~\ref{tab:llm_mix_mmlu}.
``Llama-ICL'' indicates the results using in-context learning with five examples. 
For Multitask-learning, we merge all training data from 9 fine-tune tasks (see section \ref{sec:llm_exp}) and fine-tune the Llama on this multitask datasets.
For LM-Cocktial, we use the official 5 examples to compute weights and tune a new model for each task by merging 9 fine-tuned models and the base model. 
Inspired by LoraHub~\cite{huang2023lorahub}, we also compare an alternative method to compute weight: using black-box optimization in~\cite{huang2023lorahub} to find the optimal weight assignment. We use LM-Cocktail$_{blackbox}$ to denote this variant. 
Besides, we aggregate all examples from each task to compute merging weights, and produce a unified model named LM-Cocktail$^{u}$ for all tasks, rather than generate a separate model for each task.

There are some key findings:
\begin{itemize}
    \item The performance of multitask-learning is inferior to the original llama model. This shows fine-tuning will compromise the overall generality of the original model, and also indicates there is no direct correlation between 
    these fine-tuning datasets and the tasks listed on the MMLU.
    \item LM-Cocktail achieves higher accuracy than the Llama and Llama-ICL. Despite there are no fine-tuning datasets related to MMLU tasks, our approach demonstrates a significant improvement in performance via merging fine-tuned models. LM-Cocktail merely involves recombining existing models without the need for additional model training. Furthermore, it does not introduce any latency to the inference process, while the Llama-ICL needs to process more tokens because of the added few-shot prompt.

    \item In comparison to black-box optimization, our method to compute weight is simpler yet highly effective. We observed that black-box optimization methods struggle to ensure the sum of weights equals 1, leading to suboptimal performance.

    \item The unified model LM-Cocktail$^u$ also shows superior performance, which demonstrates the proposed method is capable of simultaneously handling multiple new tasks. Besides, we further investigate the impact of the number of examples in section~\ref{sec:example_num}.
    
\end{itemize}

\subsubsection{Analysis on Encoder-based LM}
Following the setting in section~\ref{sec:unseen_llm}, we compare the performance of the original BGE model, multitask-learning model, and LM-Cocktail with different methods to mix models. We collect 9 fine-tuned models from section~\ref{sec:emb_exp}. For evaluation, we excluded tasks which has been fine-tuned in section~\ref{sec:emb_exp} (i.e.,  HotpotQA, MSMATCO, and Quora).
As reported in Table~\ref{tab:emb_mix}, LM-Cocktail achieves higher accuracy than other models. It demonstrates that we can improve the accuracy of the new task by only mixing existing language models.

\subsection{Impact of Weight $\alpha$}
In this section, we conduct a performance comparison under various weights $\alpha$. 
To eliminate the influence of other factors, we conducted experiments in the simplest configuration: merging the fine-tuned model and base model based on the weight $\alpha$. 

The results of decoders are shown in Figure~\ref{fig:alpha}, and the results of encoder models can be seen in Appendix~\ref{sec:impact_appendix}.
We incrementally varied the hyper-parameter $\alpha$ from 0 to 1, and evaluated the model's performance on the target task as well as other unrelated tasks.
It can be observed that by changing the weights of the fine-tuned model, we can significantly improve the accuracy on other tasks, even surpassing that of the base model, while ensuring that the accuracy on the target task does not decline.

% \begin{figure*}[!ht]
% \centering
% \subfigure{
% \scalebox{0.4}[0.4]{\includegraphics{figures/alpha_llm.png}}
% }
% \subfigure{
% \scalebox{0.4}[0.4]{\includegraphics{figures/alpha_emb.png}}
% }
% \caption{Performance with different mixing weight.}
% \label{fig:alpha}
% \end{figure*}

\begin{figure}[ht]
    \centering
    \scalebox{0.5}[0.5]
    {\includegraphics{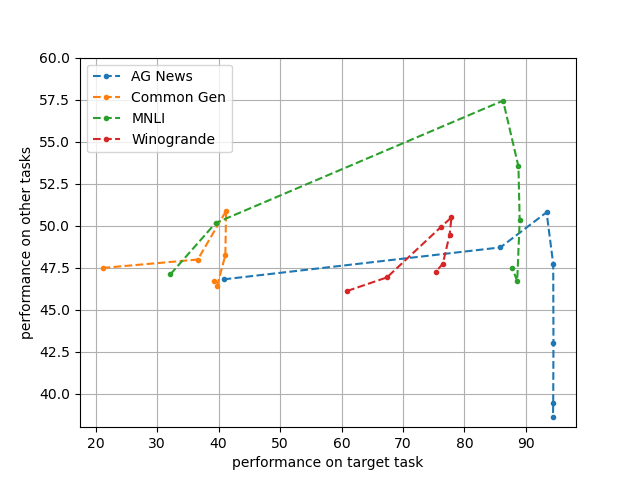}}
    \vspace{-5pt}
    \caption{Performance with different $\alpha$.}
    \label{fig:alpha}
    \vspace{-10pt}
\end{figure}

\subsection{Analysis on Number of Examples}
\label{sec:example_num}
Additionally, we investigate the effect of the number of examples. 
Given some specialist models from other tasks, LM-Cocktail needs a few examples for new task to compute the merging weights, and merge these specialist models via weighted sum. Then the merged model can be used to enhance the model fine-tuned on new task or directly perform the new task.
In this section, we evaluate the performance of merged models on new tasks following the setting in section~\ref{sec:wo_finetune}.
For decoder-based model, a total of 285 examples are provided in MMLU datasets, and we randomly sample 5, 50, and 100 examples from the entire set to merge the specialist models, and test their performance. For encoder-based model, there are a total of 115 examples, and we also sample a subset to evaluate its performance.
The average metric is reported in Table~\ref{tab:example_number}.

\begin{table}[ht]
    \centering
    \scalebox{1.0}{
    % \footnotesize
    \begin{tabular}
    {ccccc}
    \hline
Model Type & 5 & 50 & 100 & All \\
\hline
 Decoder & 47.61 & 48.13 & 48.20 & 48.21  \\ 
 % \hline
  Encoder & 48.67 & 48.76 & 48.77 & 48.73 \\
\hline
    \end{tabular}}
    \caption{The performance with different number of examples}
    \label{tab:example_number}
    \vspace{-10pt}
\end{table}

As shown in Table~\ref{tab:example_number}, 
Our approach achieves satisfactory performance using only five examples, and performance further improves with an increase in the number of examples. However, beyond fifty examples, the performance improvement becomes significantly limited.

\section{Related Work}
% In this work, we propose the model mix technology to fine-tune language models with and without labeled data, and conduct experiments on text embedding models and large language models. 
% In this section, we first introduce the embedding model and large language models, and then show the related work about model merging.
% \subsection{Embedding model}

% \subsection{Large Language Model}

\subsection{Fine-tuning of Language Model}
% Recently, large-scale pre-trained language models have demonstrated impressive zero-shot and few-shot
% learning capabilities~\cite{gpt3, bai2023qwen}. 
Fine-tuning large-scale pre-trained language models with task-specific labeled data can further enhance their corresponding abilities, which become commonplace in natural language processing~\cite{dodge2020fine}.
% In addition to fine-tuning all parameters, some Parameter-Efficient Fine-Tuning approaches have been proposed to reduce the cost~\cite{hu2021lora}.
However, catastrophic forgetting problem generally exists in the continual fine-tuning of different
language models~\cite{luo2023empirical}: 
Fine-tuning can improve the performance of the target domain, but significantly
undermine the language models’ general capabilities beyond their target domain. 
One solution is to add the data from previous tasks to maintain the previous abilities~\cite{rolnick2019experience,shin2017continual,rebuffi2017icarl}.
Some regularization-based methods also have been proposed to alleviate this problem, where the updating of model parameters is regularized to preserve the general capability of the pre-trained model\cite{kirkpatrick2017overcoming,li2017learning,rannen2017encoder}. 
Different from adding previous data, our method has no additional costs for training. Moreover,  unlike regularization-based methods, our proposed method  
requires no modification to the standard fine-tuning process.

\subsection{Model Merging}
Ensembling the outputs of many models is a popular technique for improving the accuracy 
of deep learning models~\cite{lakshminarayanan2017simple, ovadia2019can, hastie2009elements}.
However, this method requires each model
to do a separate inference, which significantly increases computational costs. 
% Especially when the number of models is large, 
% combining the outputs of multiple models is prohibitively expensive.
Instead of ensembling the outputs of models, model merging averages the weights of multiple models to improve the performance of a single model, which requires no extra computation at inference time~\cite{model-soups,ilharco2022editing}. 
Wortsman et al.~\cite {model-soups} find that
averaging the weights of models fine-tuned with different hyper-parameter configurations can often improve accuracy.
Some researchers propose more complex methods to align the parameters of different models and merge them~\cite{OTfusion_heter,merge-fisher,jin2022dataless}. 
Prateek et.al and Yu et.al propose to delete the redundant values in the delta parameter before model merging~\cite{yadav2023resolving,yu2023language}.
These methods may also be helpful for the LM-Cocktail, and we leave them for future work.

% Singh et al.~\cite{OTfusion} propose to utilize optimal transport to align neurons across
% the models before averaging their associated parameters, while Matena et al.~\cite{merge-fisher} 
% leverages the Laplace approximation by using Fisher information as the precision
% matrix for posterior distributions.
% Nguyen et al.~\cite{OTfusion_heter} introduce a cross-layer alignment method to fuse neural networks with a different number of layers.
  
% Different from these methods, we aim to preserve the models' generalization ability after fine-tuning and improve their performance on unseen tasks.

One direction relevant to our work is applying model merging in robust fine-tuning.
Wortsman et al.~\cite{wortsman2022robust} and Ilharco~\cite{open-vocab-patch} both find that the fine-tuned clip~\cite{clip} model substantially improves accuracy on a given target distribution but 
reduces robustness to distribution shifts. 
To address this problem, they use a manually set coefficient to merge the fine-tuned model and the base model.
% Croce et al.\cite{croce2023seasoning} obtain dversarially-robust model soups via linear combinations of models, which show smooth trade-off robustness on CIFAR-10 and IMAGENET datasets.
Unlike these methods, we propose to utilize not only the base pre-trained model but also the specialist models from other tasks. 
In this way, our proposed method further improves the general capabilities and even can
function in situations where fine-tuning is not feasible. Besides, we utilize a simple method to compute the merging weights for different models automatically.

The other related direction is utilizing model merging to do cross-task generalization.
Most of these methods focus on merging parameter-efficient modules (e.g., LoRA~\cite{hu2021lora}, soft prompt\cite{soft-prompt}). 
Ponti et al.\cite{ponti2023combining} introduce a latent-skill model, where they train a binary vector as router function to select skill modules for each task and then average the parameters of modules.
For the target task, Wu et al.~\cite{pi-tuning} and Lv et al.~\cite{lv-etal-2023-parameter} aggregate the parameters of lightweight task-specific experts which are learned from similar tasks.
Some works also have been proposed to merge prompt embeddings from lots of source tasks to the target domain~\cite{vu2021spot,poth2021pre,sun2023multitask}.
% However, these methods need enough data to train the router   
% function or compute the similarity between tasks.
The latest work is LoRAHub~\cite{huang2023lorahub}. Given a few examples, it uses the black-box optimization tool Shiwa~\cite{black-box} to combine multiple existing fine-tuned LoRA modules and generate a new LoRA module for the target task. 
In contrast to the above methods, we merge fine-tuned models and the base model with the aim of maintaining the general capabilities after fine-tuning. 
Meanwhile, to ensure performance on the target task, we employ losses from a small set of examples to filter out models that perform poorly on the target task.
Besides, we merge the entire model instead of the parameter-efficient modules. 

% We merge the fine-tuned model and 

% aim to maintain the general capabilities after fine-tuning on target task, while the above methods 

% we employ the loss 
% to filter the model who perform badly on target task.

% 
% In contrast to LoRAHub, we merge the entire model instead of Lora module, and aim to maintain the generalization ability after fine-tuning. 
% To 

% We use the specialist models from other tasks to boost the performance of the fine-tuned model, and propose a new approach to compute weights instead of the black-box optimization method.

\section{Conclusion}
In this work, we introduce the LM-Cocktail, a simple method to improve performance on target tasks without decreasing accuracy on other unrelated tasks. LM-Cocktail produces a resilient-tuned model by weighted averaging the parameters from different models: the model fine-tuned on the target task, the pre-trained base model, and the peer models from other domains.
The empirical results on both decoder and encoder models demonstrate that LM-Cocktail can achieve strong performance in the whole scope of general tasks while preserving a superior capacity in its targeted domain. 
We further demonstrated the effectiveness of LM-Cocktail when unable to fine-tune on domain-specific data. In such cases, our approach can merge existing models based on very few examples to enhance the accuracy of the target task.

% Entries for the entire Anthology, followed by custom entries
\bibliography{custom}
\bibliographystyle{acl_natbib}

\appendix

\section{Datasets used in fine-tuning}
\label{sec:data_appendix}
We fine-tune the Llama model on 9 different datasets,
whose details are shown in Table~\ref{tab:data_llm}.
The details of datasets used to fine-tune BGE are shown in Table~\ref{tab:data_emb}.

\begin{table}[!h]
    \centering
    % \footnotesize
    \begin{tabular}{cccc}
    \hline
        Dataset & \# train & \# test  & Metric \\
        \hline
        NQ & 30,000 & 3,610 & Exact Match \\
        SQuAD & 30,000 & 10,570 & Exact Match \\
        Hellaswag & 30,000 & 10,042 & Accuracy \\
        SST2 & 30,000 & 872 & Accuracy \\
        Winogrande & 30,000 & 1,267 & Accuracy \\
        CommenGen & 30,000 & 4,018 & ROUGE-L \\
        MRPC & 3,668 & 408 & Accuracy \\
        AG News & 30,000 & 7600 & Accuracy \\
        MNLI & 30,000 & 9,815 & Accuracy \\
    \hline
    \end{tabular}
    \caption{Statistics for the datasets used to fine-tune Llama.}
    \label{tab:data_llm}
\end{table}

\begin{table}[!h]
    \centering
    % \footnotesize
    \begin{tabular}{cccc}
    \hline
        Dataset & \# train & \# test  & Metric \\ 
        \hline
        GooAQ & 3,012,496 & -- & -- \\
        YahooAnswers & 1,198,260 & -- & -- \\
        MSMarco & 485,823 & 6,980 & NDCG@10 \\
        StackExchange & 293,951 & -- & -- \\
        ELI5 & 319,912 & -- & -- \\
        SQuAD & 86,701 & -- & -- \\
        AmazonQA & 1,095,290 & -- & -- \\
        Quora & 60,202 & 10,000 & NDCG@10 \\
        HotpotQA & 84,516 & 7,405 & NDCG@10 \\
    \hline
    \end{tabular}
    \caption{Statistics for the datasets used to fine-tune BGE.}
    \label{tab:data_emb}
\end{table}

\section{More Experimental Results}

\subsection{Detailed results for each task}
\label{sec:results_appendix}
The detailed results for each task are reported in Table~\ref{tab:appendix_base_finetune} and~\ref{tab:emb-finetune-pool-appendix}.

\subsection{Impact of $\alpha$ for encoder-based LM}
\label{sec:impact_appendix}

\begin{figure}[ht]
    \centering
    \scalebox{0.5}[0.5]
    {\includegraphics{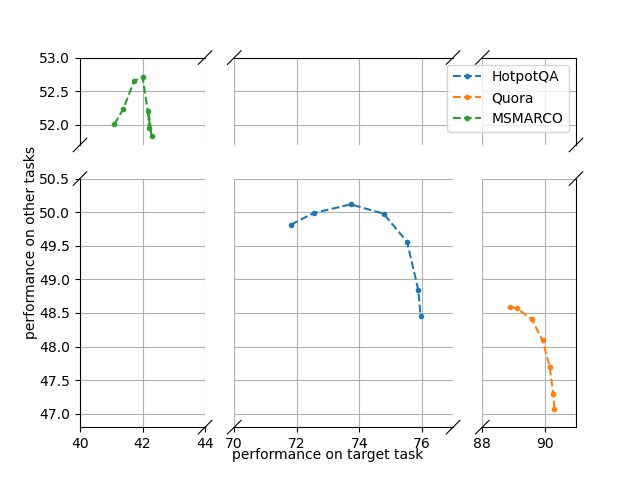}}
    \caption{Performance of encoder-based LMs with different merging weights.}
    \label{fig:alpha-emb}
    \vspace{-10pt}
\end{figure}

The performance of encoder-based LMs with different merging weight are shown in Figure~\ref{fig:alpha-emb}.

\begin{sidewaystable*}
\centering
\caption{The detailed results of base model, fine-tuned model, and resilient-tuned model via LM-Cocktail. We use the name of task to denote the model fine-tuned on this task (e.g., AG News in the first row represent the model fine-tuned on AG News dataset).  LM-Cocktail$_2$ is produced by merging the base model and fine-tuned model, while LM-Cocktail$_{10}$
merges fine-tuned model, base model, and other models fine-tuned on 8 different tasks. }
\label{tab: more_finetune_base}
\resizebox{1.00\textwidth}{!}
{
% \setlength{\tabcolsep}{0.9mm}
% {
\begin{tabular}{|c|c|ccc|ccc|ccc|ccc|ccc|ccc|ccc|ccc|ccc|}
\hline
Dataset &Llama & AG News & LM-Cocktail$_2$ & LM-Cocktail$_{10}$ & Commen Gen & LM-Cocktail$_2$ & LM-Cocktail$_{10}$ & SQuAD & LM-Cocktail$_2$ & LM-Cocktail$_{10}$ & NQ & LM-Cocktail$_2$ & LM-Cocktail$_{10}$ & Winogrande & LM-Cocktail$_2$ & LM-Cocktail$_{10}$ & Hellaswag & LM-Cocktail$_2$ & LM-Cocktail$_{10}$ & MNLI & LM-Cocktail$_2$ & LM-Cocktail$_{10}$ & MRPC & LM-Cocktail$_2$ & LM-Cocktail$_{10}$ & SST2 & LM-Cocktail$_2$ & LM-Cocktail$_{10}$ \\ 
\hline 
arc-c & 36.05 & 34.68 & 36.91 & 38.37 & 37.68 & 37.77 & 42.15 & 36.05 & 42.32 & 43.43 & 41.46 & 43.78 & 43.52 & 35.88 & 38.8 & 41.72 & 38.54 & 37.68 & 41.12 & 33.48 & 35.79 & 37.17 & 30.73 & 32.1 & 35.79 & 32.96 & 37.68 & 38.2 \\ 
arc-e & 51.59 & 41.65 & 51.08 & 54.16 & 60.21 & 60.47 & 66.51 & 58.39 & 65.58 & 68.96 & 68.84 & 68.92 & 71.71 & 55.56 & 60.8 & 66.05 & 60.0 & 59.87 & 65.37 & 40.21 & 49.18 & 52.52 & 34.97 & 42.11 & 48.37 & 45.41 & 50.06 & 52.18 \\ 
natural-questions & 0.0 & 0.0 & 0.03 & 0.03 & 0.0 & 0.11 & 6.84 & 14.07 & 14.79 & 16.12 & 29.09 & 29.25 & 29.64 & 0.66 & 0.91 & 9.81 & 0.06 & 0.53 & 9.67 & 0.0 & 0.0 & 0.0 & 0.03 & 0.03 & 0.03 & 0.0 & 0.0 & 0.0 \\ 
% Avg. CQA & 29.21 & 25.44 & 29.34 & 30.85 & 32.63 & 32.78 & 38.5 & 36.17 & 40.9 & 42.84 & 46.46 & 47.32 & 48.29 & 30.7 & 33.51 & 39.19 & 32.87 & 32.69 & 38.72 & 24.56 & 28.32 & 29.89 & 21.91 & 24.75 & 28.06 & 26.12 & 29.25 & 30.13 \\ 
\hline 
copa & 69.0 & 57.0 & 70.0 & 67.0 & 64.0 & 66.0 & 78.0 & 65.0 & 69.0 & 76.0 & 70.0 & 76.0 & 75.0 & 74.0 & 76.0 & 77.0 & 69.0 & 71.0 & 77.0 & 59.0 & 71.0 & 74.0 & 47.0 & 63.0 & 63.0 & 64.0 & 72.0 & 72.0 \\ 
hellaswag & 71.58 & 65.76 & 71.8 & 72.1 & 67.73 & 72.02 & 73.67 & 61.0 & 68.25 & 69.92 & 69.01 & 71.57 & 72.22 & 64.8 & 70.48 & 71.64 & 77.2 & 79.0 & 78.61 & 63.31 & 71.8 & 71.42 & 54.09 & 68.69 & 69.97 & 65.29 & 69.57 & 70.3 \\ 
piqa & 75.73 & 73.07 & 76.17 & 75.68 & 74.65 & 76.77 & 76.71 & 74.54 & 76.12 & 75.95 & 76.44 & 76.5 & 76.33 & 74.21 & 76.77 & 76.71 & 75.95 & 77.97 & 77.2 & 72.14 & 76.5 & 75.84 & 69.97 & 75.73 & 75.68 & 74.48 & 76.71 & 75.84 \\ 
% Avg. Commonsense & 72.1 & 65.28 & 72.66 & 71.59 & 68.79 & 71.6 & 76.13 & 66.85 & 71.12 & 73.96 & 71.82 & 74.69 & 74.52 & 71.0 & 74.42 & 75.12 & 74.05 & 75.99 & 77.6 & 64.82 & 73.1 & 73.75 & 57.02 & 69.14 & 69.55 & 67.92 & 72.76 & 72.72 \\ 
\hline 
winogrande & 60.93 & 53.59 & 58.88 & 63.3 & 58.17 & 59.67 & 65.82 & 57.22 & 61.4 & 65.9 & 59.98 & 60.77 & 67.4 & 75.45 & 77.9 & 77.03 & 62.98 & 64.48 & 66.85 & 61.8 & 62.67 & 64.56 & 52.01 & 56.27 & 59.59 & 59.67 & 61.33 & 64.4 \\ 
wsc & 60.58 & 63.46 & 63.46 & 63.46 & 63.46 & 61.54 & 63.46 & 63.46 & 63.46 & 63.46 & 57.69 & 60.58 & 62.5 & 58.65 & 53.85 & 62.5 & 60.58 & 63.46 & 63.46 & 63.46 & 60.58 & 61.54 & 64.42 & 63.46 & 63.46 & 63.46 & 63.46 & 63.46 \\ 
wsc273 & 75.82 & 72.53 & 75.82 & 78.75 & 64.84 & 74.73 & 79.85 & 69.23 & 76.19 & 78.39 & 72.16 & 78.02 & 76.92 & 79.85 & 84.25 & 84.98 & 76.19 & 79.49 & 81.68 & 74.73 & 78.75 & 78.75 & 58.61 & 67.03 & 72.89 & 72.89 & 76.92 & 78.75 \\ 
% Avg. Coreference & 65.78 & 63.19 & 66.05 & 68.51 & 62.16 & 65.31 & 69.71 & 63.3 & 67.02 & 69.25 & 63.28 & 66.46 & 68.94 & 71.32 & 72.0 & 74.84 & 66.58 & 69.14 & 70.67 & 66.66 & 67.33 & 68.28 & 58.35 & 62.26 & 65.31 & 65.34 & 67.24 & 68.87 \\ 
\hline 
mrpc & 31.86 & 31.62 & 31.62 & 31.86 & 32.11 & 31.62 & 31.62 & 31.62 & 31.86 & 32.84 & 31.62 & 31.86 & 31.62 & 34.56 & 32.35 & 32.11 & 31.62 & 31.62 & 31.62 & 74.02 & 73.04 & 70.1 & 85.78 & 73.77 & 80.88 & 31.62 & 31.62 & 31.62 \\ 
paws & 55.96 & 55.76 & 55.88 & 55.8 & 55.74 & 55.64 & 56.07 & 55.43 & 55.96 & 55.73 & 55.93 & 55.69 & 57.06 & 55.7 & 56.01 & 56.91 & 55.97 & 55.97 & 55.83 & 66.77 & 60.6 & 66.26 & 44.75 & 47.44 & 52.02 & 55.8 & 55.8 & 55.86 \\ 
qqp & 69.43 & 65.29 & 66.07 & 64.28 & 61.56 & 68.07 & 66.41 & 63.13 & 63.66 & 63.87 & 65.5 & 67.18 & 68.04 & 61.02 & 69.38 & 69.5 & 63.13 & 66.12 & 63.92 & 80.61 & 76.03 & 81.36 & 49.76 & 53.94 & 68.58 & 63.19 & 63.19 & 63.19 \\ 
% Avg. Paraphrase & 52.42 & 50.89 & 51.19 & 50.65 & 49.8 & 51.77 & 51.37 & 50.06 & 50.5 & 50.81 & 51.01 & 51.58 & 52.24 & 50.43 & 52.58 & 52.84 & 50.24 & 51.24 & 50.45 & 73.8 & 69.89 & 72.57 & 60.1 & 58.38 & 67.16 & 50.2 & 50.2 & 50.22 \\ 
\hline 
rte & 70.04 & 51.26 & 74.73 & 75.09 & 48.01 & 50.54 & 68.59 & 75.45 & 79.78 & 79.78 & 54.15 & 71.48 & 68.95 & 47.65 & 53.79 & 75.81 & 47.65 & 48.01 & 77.98 & 81.95 & 76.17 & 77.62 & 58.12 & 49.1 & 57.4 & 47.29 & 47.65 & 50.54 \\ 
snli & 34.02 & 33.44 & 50.22 & 37.34 & 37.45 & 35.94 & 45.01 & 44.51 & 33.04 & 33.34 & 41.05 & 37.44 & 62.2 & 36.99 & 40.03 & 48.67 & 41.08 & 43.99 & 49.27 & 84.95 & 87.56 & 88.82 & 35.06 & 37.17 & 43.8 & 32.95 & 32.95 & 32.95 \\ 
mnli-m & 32.14 & 34.28 & 51.6 & 41.87 & 31.29 & 31.82 & 52.33 & 39.27 & 33.38 & 37.11 & 38.4 & 35.16 & 47.5 & 34.16 & 33.21 & 50.26 & 35.51 & 35.74 & 54.29 & 87.9 & 88.88 & 89.23 & 39.96 & 40.53 & 46.52 & 32.74 & 32.74 & 32.75 \\ 
mnli-mm & 32.01 & 35.43 & 52.15 & 43.93 & 31.76 & 32.36 & 51.99 & 38.3 & 33.59 & 37.08 & 38.06 & 34.46 & 47.41 & 33.78 & 33.26 & 50.03 & 35.46 & 35.74 & 54.12 & 87.93 & 89.52 & 89.7 & 39.3 & 40.22 & 51.06 & 32.95 & 32.95 & 33.05 \\ 
qnli & 55.19 & 52.44 & 64.21 & 57.79 & 54.7 & 57.81 & 73.9 & 50.61 & 54.18 & 51.78 & 53.91 & 61.07 & 68.79 & 54.9 & 60.31 & 71.13 & 61.43 & 63.54 & 76.26 & 79.04 & 80.18 & 82.3 & 69.49 & 56.82 & 56.76 & 50.54 & 50.63 & 50.56 \\ 
% Avg. NLI & 44.68 & 41.37 & 58.58 & 51.2 & 40.64 & 41.69 & 58.36 & 49.63 & 46.8 & 47.82 & 45.11 & 47.92 & 58.97 & 41.5 & 44.12 & 59.18 & 44.23 & 45.41 & 62.38 & 84.35 & 84.46 & 85.53 & 48.38 & 44.77 & 51.11 & 39.29 & 39.38 & 39.97 \\ 
\hline 
multirc & 44.46 & 26.02 & 56.15 & 56.71 & 43.58 & 38.27 & 44.65 & 7.77 & 27.13 & 53.16 & 53.93 & 49.31 & 59.99 & 60.09 & 61.58 & 60.38 & 26.78 & 30.63 & 38.0 & 62.77 & 74.32 & 76.42 & 51.81 & 58.23 & 47.13 & 0.0 & 36.81 & 49.52 \\ 
openbookqa & 47.0 & 44.0 & 44.8 & 46.0 & 47.6 & 49.0 & 50.0 & 45.2 & 48.0 & 50.4 & 51.4 & 51.6 & 52.6 & 49.6 & 51.8 & 52.4 & 48.0 & 49.8 & 51.0 & 45.0 & 45.4 & 45.2 & 41.0 & 43.0 & 44.2 & 44.0 & 43.4 & 45.2 \\ 
boolq & 75.26 & 59.42 & 76.67 & 76.18 & 69.11 & 72.39 & 72.08 & 67.58 & 77.74 & 77.77 & 76.06 & 77.16 & 78.5 & 79.66 & 79.51 & 78.75 & 65.78 & 65.11 & 67.09 & 60.4 & 78.13 & 78.1 & 43.55 & 55.14 & 42.66 & 37.83 & 41.96 & 53.06 \\ 
squad-v1 & 0.06 & 0.04 & 0.06 & 1.26 & 16.41 & 28.94 & 73.15 & 86.77 & 85.67 & 86.94 & 54.61 & 58.86 & 77.4 & 28.86 & 29.22 & 72.74 & 2.88 & 10.47 & 70.06 & 0.03 & 0.16 & 7.28 & 0.07 & 0.03 & 0.22 & 0.03 & 0.79 & 16.36 \\ 
% Avg. ReadingComp & 41.7 & 32.37 & 44.42 & 45.04 & 44.18 & 47.15 & 59.97 & 51.83 & 59.63 & 67.07 & 59.0 & 59.23 & 67.12 & 54.55 & 55.53 & 66.07 & 35.86 & 39.0 & 56.54 & 42.05 & 49.5 & 51.75 & 34.1 & 39.1 & 33.55 & 20.46 & 30.74 & 41.03 \\ 
\hline 
sentiment140 & 66.85 & 56.27 & 72.7 & 81.62 & 77.99 & 84.4 & 93.59 & 83.84 & 84.96 & 91.09 & 91.36 & 89.69 & 93.59 & 63.23 & 72.7 & 90.53 & 73.82 & 84.68 & 93.59 & 48.75 & 89.97 & 92.76 & 49.58 & 54.32 & 56.82 & 92.2 & 87.47 & 92.48 \\ 
sst2 & 63.3 & 60.21 & 86.58 & 87.04 & 62.73 & 78.44 & 93.46 & 74.43 & 65.14 & 86.93 & 66.63 & 79.36 & 92.78 & 60.55 & 76.72 & 89.45 & 54.47 & 80.96 & 93.92 & 51.95 & 71.56 & 86.93 & 48.85 & 50.92 & 50.92 & 95.53 & 96.56 & 96.56 \\ 
yelp & 82.9 & 51.58 & 95.96 & 96.88 & 94.95 & 93.85 & 97.75 & 96.96 & 96.24 & 96.89 & 81.87 & 83.55 & 96.88 & 84.93 & 93.2 & 96.73 & 87.89 & 90.95 & 97.84 & 52.7 & 85.22 & 96.79 & 48.0 & 52.39 & 68.63 & 96.4 & 83.29 & 92.8 \\ 
% Avg. Sentiment & 71.02 & 56.02 & 85.08 & 88.51 & 78.56 & 85.56 & 94.94 & 85.08 & 82.11 & 91.64 & 79.95 & 84.2 & 94.42 & 69.57 & 80.87 & 92.24 & 72.06 & 85.53 & 95.12 & 51.13 & 82.25 & 92.16 & 48.81 & 52.54 & 58.79 & 94.71 & 89.1 & 93.95 \\ 
\hline 
common-gen & 21.14 & 0.0 & 0.0 & 0.63 & 39.2 & 41.22 & 41.45 & 20.35 & 35.82 & 36.0 & 30.74 & 33.43 & 32.9 & 28.89 & 29.83 & 36.11 & 34.62 & 31.85 & 39.28 & 0.0 & 0.0 & 0.0 & 0.0 & 0.0 & 0.0 & 0.0 & 0.0 & 15.04 \\ 
e2e-nlg & 34.41 & 0.0 & 0.0 & 33.56 & 27.17 & 48.77 & 48.23 & 21.73 & 44.29 & 43.34 & 16.29 & 36.68 & 39.0 & 31.26 & 44.01 & 46.14 & 38.77 & 39.98 & 43.9 & 0.01 & 0.39 & 9.64 & 0.04 & 0.04 & 0.04 & 0.0 & 24.01 & 39.34 \\ 
dart & 32.58 & 0.1 & 0.16 & 0.22 & 29.03 & 40.89 & 35.98 & 20.73 & 24.63 & 21.91 & 16.85 & 35.85 & 33.3 & 31.06 & 37.34 & 38.86 & 28.54 & 32.88 & 41.23 & 0.02 & 0.02 & 0.05 & 0.05 & 0.05 & 0.05 & 0.0 & 0.33 & 0.15 \\ 
% Avg. Data2Text & 29.38 & 0.03 & 0.05 & 11.47 & 31.8 & 43.63 & 41.89 & 20.94 & 34.91 & 33.75 & 21.29 & 35.32 & 35.07 & 30.41 & 37.06 & 40.37 & 33.98 & 34.9 & 41.47 & 0.01 & 0.14 & 3.23 & 0.03 & 0.03 & 0.03 & 0.0 & 8.12 & 18.18 \\ 
\hline 
aeslc & 3.51 & 0.02 & 0.16 & 0.42 & 2.65 & 4.01 & 1.48 & 4.58 & 4.36 & 3.78 & 4.28 & 4.52 & 1.36 & 3.41 & 4.21 & 1.89 & 2.18 & 2.79 & 1.84 & 0.0 & 0.69 & 0.78 & 0.03 & 0.01 & 0.07 & 0.0 & 1.58 & 0.79 \\ 
ag-news & 40.8 & 94.42 & 94.46 & 94.41 & 39.12 & 50.33 & 88.04 & 75.11 & 52.53 & 54.53 & 70.29 & 47.32 & 87.24 & 38.18 & 39.14 & 82.71 & 49.92 & 56.34 & 87.68 & 32.14 & 53.54 & 66.67 & 25.8 & 39.61 & 64.68 & 33.7 & 31.95 & 35.33 \\ 
gigaword & 3.8 & 0.0 & 0.16 & 0.09 & 6.51 & 13.28 & 1.3 & 19.32 & 14.15 & 3.16 & 4.94 & 5.13 & 5.52 & 10.4 & 5.7 & 5.64 & 7.98 & 7.91 & 7.16 & 0.0 & 3.58 & 0.52 & 0.0 & 0.0 & 0.0 & 0.0 & 0.44 & 0.14 \\ 
% Avg. Summarize & 16.04 & 31.48 & 31.6 & 31.64 & 16.09 & 22.54 & 30.27 & 33.0 & 23.68 & 20.49 & 26.5 & 18.99 & 31.37 & 17.33 & 16.35 & 30.08 & 20.03 & 22.35 & 32.23 & 10.71 & 19.27 & 22.66 & 8.61 & 13.21 & 21.58 & 11.23 & 11.32 & 12.08 \\ 
\hline 
Avg & 46.6 & 40.44 & 49.28 & 49.86 & 46.65 & 50.56 & 58.0 & 50.72 & 52.77 & 55.19 & 51.42 & 53.74 & 59.26 & 48.27 & 51.44 & 59.14 & 47.13 & 49.95 & 58.56 & 48.84 & 54.71 & 57.41 & 38.09 & 40.7 & 44.04 & 40.83 & 43.46 & 46.75 \\ 
\hline 
\label{tab:appendix_base_finetune}
\end{tabular}
}
% }
\end{sidewaystable*}

\begin{table*}[!htbp]
    \centering
    \scalebox{0.7}{
    \footnotesize
    \begin{tabular}
    {|c|c|ccc|ccc|ccc|}
    \hline
Dataset & BGE & Hotpotqa & LM-Cocktail$_2$ & LM-Cocktail$_{10}$ & Quora & LM-Cocktail$_2$ &  LM-Cocktail$_{10}$ & MSMARCO & LM-Cocktail$_2$ &  LM-Cocktail$_{10}$ \\ 
\hline
ArguAna & 63.61 & 59.11 & 62.1 & 63.17 & 62.69 & 63.02 & 63.92 & 60.76 & 62.82 & 63.32 \\ 
ClimateFEVER & 29.51 & 28.45 & 30.04 & 30.4 & 26.61 & 28.8 & 29.09 & 27.82 & 29.34 & 29.28 \\ 
DBPedia & 40.56 & 40.15 & 41.56 & 42.11 & 38.68 & 39.86 & 40.17 & 40.38 & 41.48 & 41.59 \\ 
FEVER & 83.66 & 84.89 & 85.63 & 86.7 & 82.36 & 83.42 & 85.07 & 84.23 & 85.34 & 85.97 \\ 
FiQA2018 & 39.11 & 38.72 & 39.68 & 41.28 & 36.6 & 38.74 & 40.91 & 38.88 & 39.73 & 41.51 \\ 
HotpotQA & 71.81 & 75.96 & 74.78 & 74.67 & 67.94 & 70.61 & 70.9 & 69.06 & 71.41 & 71.42 \\ 
MSMARCO & 41.15 & 38.98 & 40.64 & 40.66 & 40.1 & 40.96 & 40.96 & 42.23 & 42.01 & 41.88 \\ 
NFCorpus & 36.83 & 36.34 & 37.27 & 37.54 & 37.09 & 37.18 & 37.52 & 37.56 & 37.45 & 37.59 \\ 
NQ & 51.05 & 47.88 & 51.36 & 52.28 & 50.01 & 51.1 & 52.58 & 53.37 & 53.66 & 54.42 \\ 
QuoraRetrieval & 88.9 & 88.2 & 88.71 & 88.77 & 90.31 & 89.93 & 89.81 & 88.6 & 88.9 & 89.01 \\ 
SCIDOCS & 21.48 & 19.81 & 21.1 & 21.65 & 19.97 & 20.77 & 21.42 & 21.37 & 21.53 & 21.86 \\ 
SciFact & 73.81 & 75.08 & 74.57 & 75.53 & 74.72 & 74.78 & 75.13 & 73.58 & 74.2 & 74.38 \\ 
Touche2020 & 19.54 & 19.04 & 19.53 & 20.12 & 19.53 & 18.95 & 19.8 & 22.08 & 21.31 & 21.13 \\ 
TRECCOVID & 67.18 & 61.13 & 66.15 & 66.36 & 61.5 & 63.31 & 67.3 & 71.33 & 70.08 & 71.53 \\ 
CQADupstack & 41.04 & 40.54 & 41.27 & 42.29 & 41.33 & 41.71 & 42.7 & 38.67 & 40.78 & 41.99 \\ 
\hline
Avg & 51.28 & 50.29 & 51.63 & 52.24 & 49.96 & 50.88 & 51.82 & 51.33 & 52.0 & 52.46 \\ 
\hline 
    \end{tabular}}
    \caption{The detailed results of base model, fine-tuned model, and resilient-tuned model via LM-Cocktail. We use the name of task to denote the model fine-tuned on this task (e.g., HotpotQA in the first row represent the model fine-tuned on HotpotQA dataset). 
    LM-Cocktail$_2$ is produced by merging the base model and fine-tuned model, while LM-Cocktail$_{10}$
merges fine-tuned model, base model, and other models fine-tuned on 8 different tasks. }
    \label{tab:emb-finetune-pool-appendix}
    \vspace{-10pt}
\end{table*}

\end{document}